# Adaptive Entangled Neurons in the Brain and Natural Intelligence Simulations for Artificial General Intelligence

Haochen Li[1,2], Xinshuai Guo[3], Jingdong Ouyang[4], Wei Zhang[5], Leilei Shi[3,5,*]

[1] Laboratory for Artificial Intelligence in Design, Hong Kong S.A.R., China
[2] School of Design, Royal College of Art, London, SW7 2EU, United Kingdom
[3] University of Science and Technology of China (USTC), School of Management, Hefei 230026, P. R. China
[4] Red Horse Investments Group, Beijing 100032, China
[5] Beijing Shangdafei Science & Technology Co., Ltd, Beijing 101300, P. R. China
*Corresponding author's E-mails: Shileilei8@163.com (LS)

## ABSTRACT

Understanding the brain's mechanisms is crucial in neuroscience, particularly for simulating natural intelligence and advancing artificial general intelligence (AGI). Liu et al. (2024) and Deco et al. (2025) propose that quantum entanglement may underlie the synchronization of millions of neurons, introducing the Liu-Chen-Ao (LCA) hypothesis of entangled nerve fibers. However, directly testing this idea remains challenging. Here we employ a nonlocal many-body wave equation, introducing a generalized probability-wave framework to model collective behaviors among adaptive, interacting agents, particles, or neurons. This approach enables testable eigenmodes, offering an indirect means of examining the LCA hypothesis by observing patterns in collective trading behaviors. These patterns, rooted in behavioral psychology, are thought to mirror the internal brain's decision-making and underlying mechanisms. Our analysis of Chinese intraday stock market data shows that adaptive entangled game modes explain 82%-94% of decision patterns, sharply contrasting neoclassical finance predictions assuming independent rational agents. Additionally, 2–12% of actions adapt to news, events, and environmental changes, marked by rapid equilibrium shifts, while truly independent behaviors make up less than 5%. These results empirically support the LCA hypothesis and underscore the need to simulate the brain's mechanisms and generate adaptive, entangled modules for AGI. They also expose the shortcomings of conventional artificial neural networks (ANNs) that depend on vast, opaque parameters. By combining ANN-based AI with probability-wave-based entangled brain simulations, machine learning enables the creation of human-like processing units (HPUs). Such HPUs promise more efficient and robust AGI, particularly for embodied intelligence and robotics.

# 1. Introduction

"I think the next century (the 21st century) will be the century of complexity." –Stephen Hawking [1]

Artificial intelligence (AI) is reshaping society and accelerating scientific discovery, establishing itself as an integral part of modern science [2]. At the scientific frontier, quantum machine learning is exploring how quantum information resources, such as entangled states, can enhance learning and inference [3, 4]. Despite rapid progress, however, a comprehensive and interpretable theory of artificial general intelligence (AGI) remains elusive [5-7]. While there is no universally accepted definition of AGI, basic consensus has emerged: AGI is "a system that can do almost all cognitive tasks that a human can do" [8-10]. Achieving this is daunting due to the multidimensional complexity of natural intelligence simulations, intricate neural interactions, and our incomplete mechanistic understanding. The advent of possible AGI models, such as foundation models (FMs) [11, 12], parallel intelligence frameworks [13], and ensemble learning approaches [14], further underscores these challenges. An FM is any model that is trained on broad data (generally using self-supervision at scale) that can be adapted (e.g., fine-tuned) to a wide range of downstream tasks. Yet we lack a clear understanding of how it works, when it fails, and what it is even capable of because of its emergent properties.

To address these obstacles, we apply a trading volume-price probability wave equation in finance [15-17] and a nonlocal many-body probability-wave equation in complex quantum systems [18], presenting a generalized natural intelligence probability-wave-based framework for modeling the collective behaviors of adaptive, interacting particles [19-22], agents, and neurons in complex adaptive systems [23-26]. This nonlocal approach yields testable eigenmodes and provides an indirect means of examining the Liu-Chen-Ao (LCA) hypothesis of entangled nerve fibers in the brain that needs to perform time-critical computations to ensure survival [27-29] by analyzing collective trading behaviors. Such observable behaviors, rooted in behavioral psychology [30, 31][①], are believed to reflect internal decision-making processes and the underlying brain mechanisms associated with natural intelligence. This interdisciplinary study can help measure human likeness of AI to natural intelligence [32] and simulate human brain mechanisms, constructing behavioral, entangled brain-like processing units to ultimately achieve AGI architectures. This is the gap left by conventional artificial neural network (ANN)-based AI, which relies on trillions of opaque parameters, a challenge into which theoretical biophysicists and behavioral econophysicists enable valuable insights (see Fig. 1).

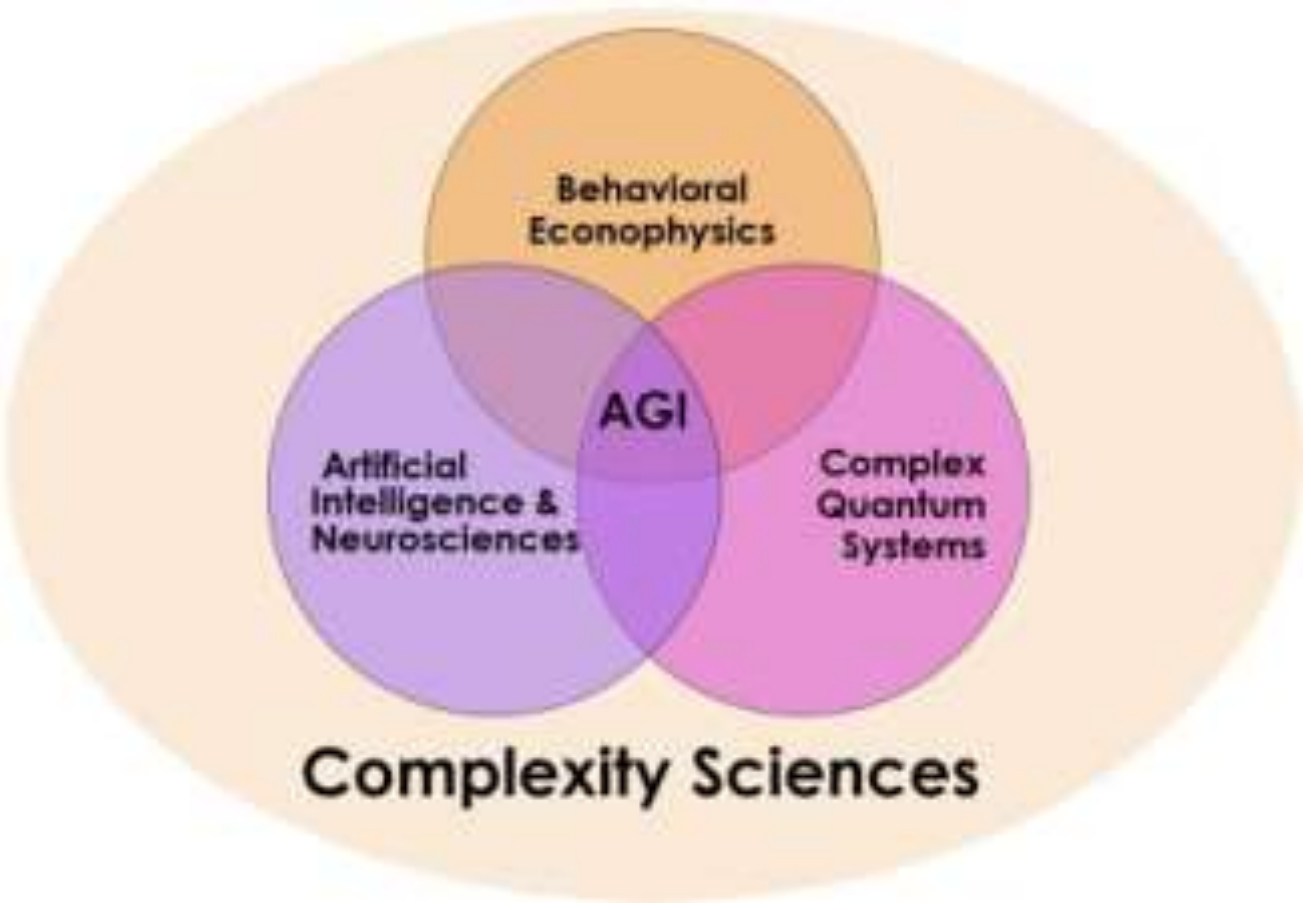

Fig. 1. The probability-wave paradigm for AGI bridges behavioral econophysics, artificial intelligence & neurosciences, and nonlocal many-body entangled systems under the unified lens of complex adaptive systems.

AGI is a central system of simulating natural intelligence and must be reliable, useful, and adjustable AI capable of accomplishing a wide range of complex downstream tasks. We argue that true AGI should have some defined functional boundaries and can operate as a multimodal simulation system in open, dynamic environments, consolidating

① Home of the Association for Behavior Analysis International (ABAI): https://www.abainternational.org/welcome.aspx

methodologies for building machine learning systems across a wide range of applications. It must perform a wide range of human-like cognitive and complex tasks, adapt responsively to sensitive or reinforcement variables, and generalize across task transfers in evolutionary processes [10, 33, 34], integrating modules capable of performing nonlocal, adaptive, entangled game behaviors in machine learning.

Adaptive entangled games specify mutually influential, interaction-coherent entangled games between inseparable, opposite agents, such as buyers and sellers in stock markets, who adapt to sensitive environmental variables within reinforcement learning feedback loops. Mathematically, such adaptive entangled game dynamics generate a squared zero-order Bessel volume distribution (see Fig. 2a). The sensitive variables, to which agents, neurons or nodes respond, may range from economic factors (e.g., price, wealth) to physical quantities (e. g., temperature, frequency), and physiological reinforcements (e. g., foods, protein, visual light, gravity fields), etc.

Since the emergence of econophysics in the 1990s, human behavioral models have advanced our understanding of economics, behavioral finance, and social systems [15-17, 35-43]. They have also propelled AI development, most notably in the evolution of ANNs powering large language models (LLMs), such as OpenAI's LLMs and DeepSeek's reinforcement learning (RL)-based LLMs [44-47]. Analytical behavioral modeling thus provides a promising approach to elucidating AGI mechanisms.

Historically, AI has developed along three paradigms: symbolism (rule-based systems), connectionism (ANNs and LLMs), and behaviorism (intelligence decision-making and reinforcement learning inspired by psychology) [48]. While symbolism and connectionism have delivered impressive results—e.g., AlphaGo [49], transformer-based models for financial forecasting [42]—they are data-driven, often require immense computational resources, and lack interpretability. In contrast, behaviorist approaches attempts to offer interpretable mechanisms in AI systems.

Given the complexity of human-like intelligence, we decompose AGI into modular simulation units, focusing here on adaptive entangled games within a nonlocal probability-wave framework. Probability waves, first introduced in quantum mechanics in 1926 [50, 51], capture uncertainty and steerability—key features of intelligent behavior. Unlike classical waves, its intensity is defined by probability rather than amplitude. In 2006, Shi [15] applied this concept to model intraday trading volume distributions (see Fig. 2a), and later Shi et al. [18] extended this trading-volume price probability wave theory to explain nonlocal coherent entanglement in many-body quantum systems [19-22] (Fig. 2b), thereby providing new interpretations of phenomena such as the Einstein–Podolsky–Rosen (EPR) paradox. This novel theory finds existence of two orthogonal eigenstates: independent, quantized energy eigenstates and interaction-coherent frequency eigenstates in a unified framework. The frequency eigenstates are not quantized, determined by eigenvalues in interaction wave functions. The theory explains strong correlations of bipartite frequencies/colors in a coherent entangled state through conservation of interaction-coherent eigenfrequencies, while it needs further experimental falsification. Many-body entangled states exist in interaction-coherent frequency eigenstates rather than superposition of energy eigenstates (see Fig. 2c).

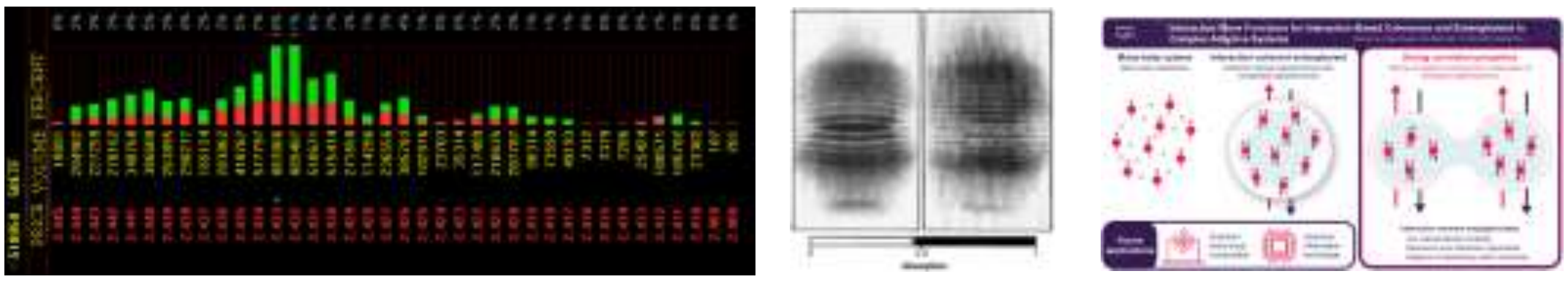

(a) (b) (c)

Fig.2. Interaction-coherent entangled states in stock markets and complex quantum systems

Note: (a) Coherent entangled game patterns in stock markets—mutually influential traders adapt to buy low and sell high, yielding intraday, nonlinear cumulative trading volume distributions across price ranges; (b) Interaction-coherent entanglement patterns in a Bose–Einstein condensate (BEC) [21]; (c) Conservation of interaction-coherent eigenfrequencies explains the strong correlation of bipartite colors in coherent entangled states.

Probability, as a unifying measure, links preferences in economics, operant frequencies in behavior analysis, uncertainties in quantum mechanics, and intelligence behaviors in AI. Our generalized probability-wave-based framework for AGI systems integrates insights across these domains. According to behavioral psychology, observable trading behaviors can serve as proxies for internal brain processes and intelligence decision-making in a three-term contingency (Fig. 3) [30, 31], a principle we leverage to test the LCA hypothesis where a potential solution lies in the nonlocal, distributed computation at the whole-brain level using tick-by-tick high-frequency data from Chinese stock markets.

To manage the high dimensionality of generalized natural intelligence, we utilize Skinner-Shi coordinates (see Fig. 4) [15, 30], reducing the system to a two-dimensional framework that introduces concepts such as crowd operant momentum, operant force, and operant energy, and links them to decision-making dynamics.

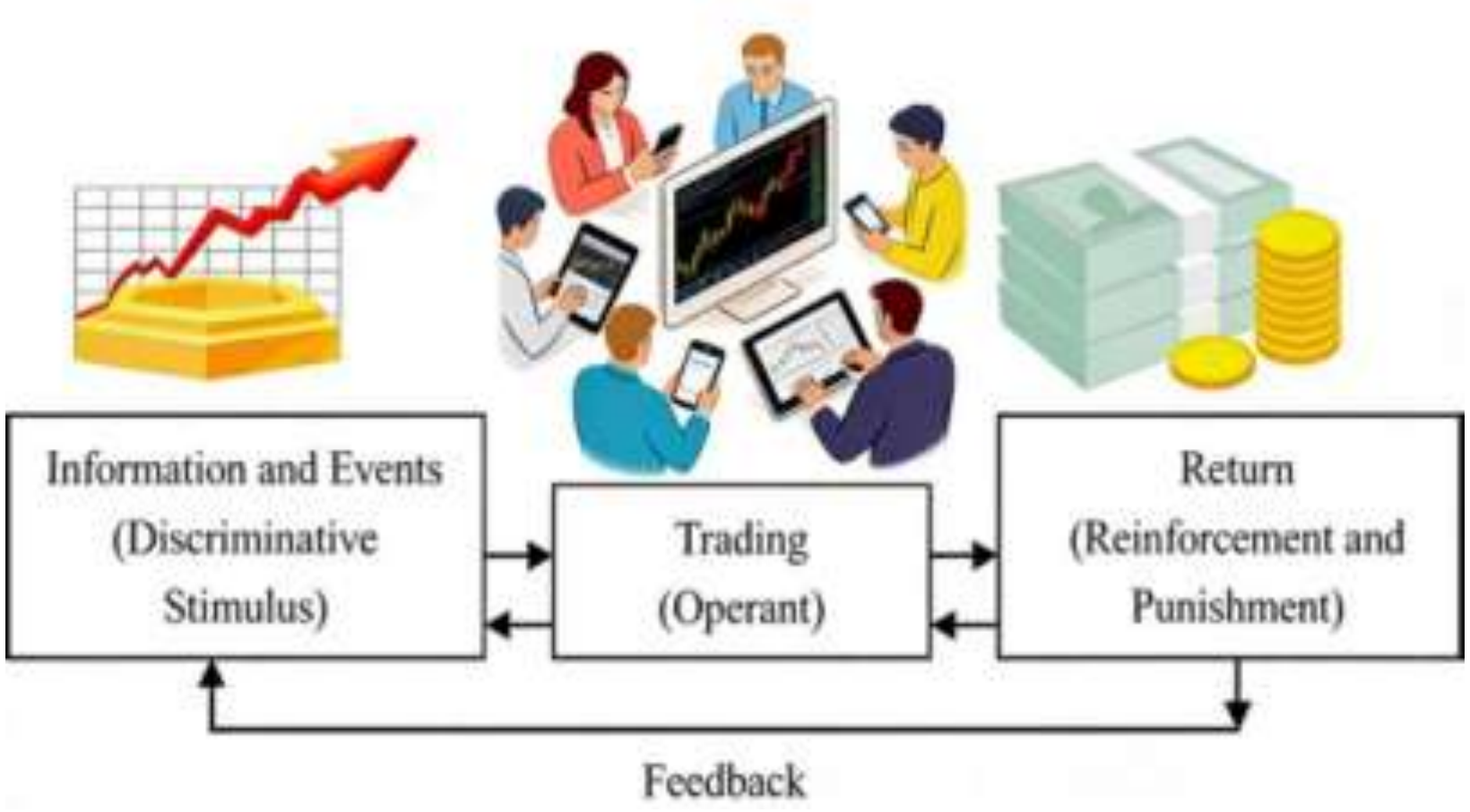

Fig. 3. Mechanisms of adaptive entangled games between buyers and sellers in a three-term contingency framework

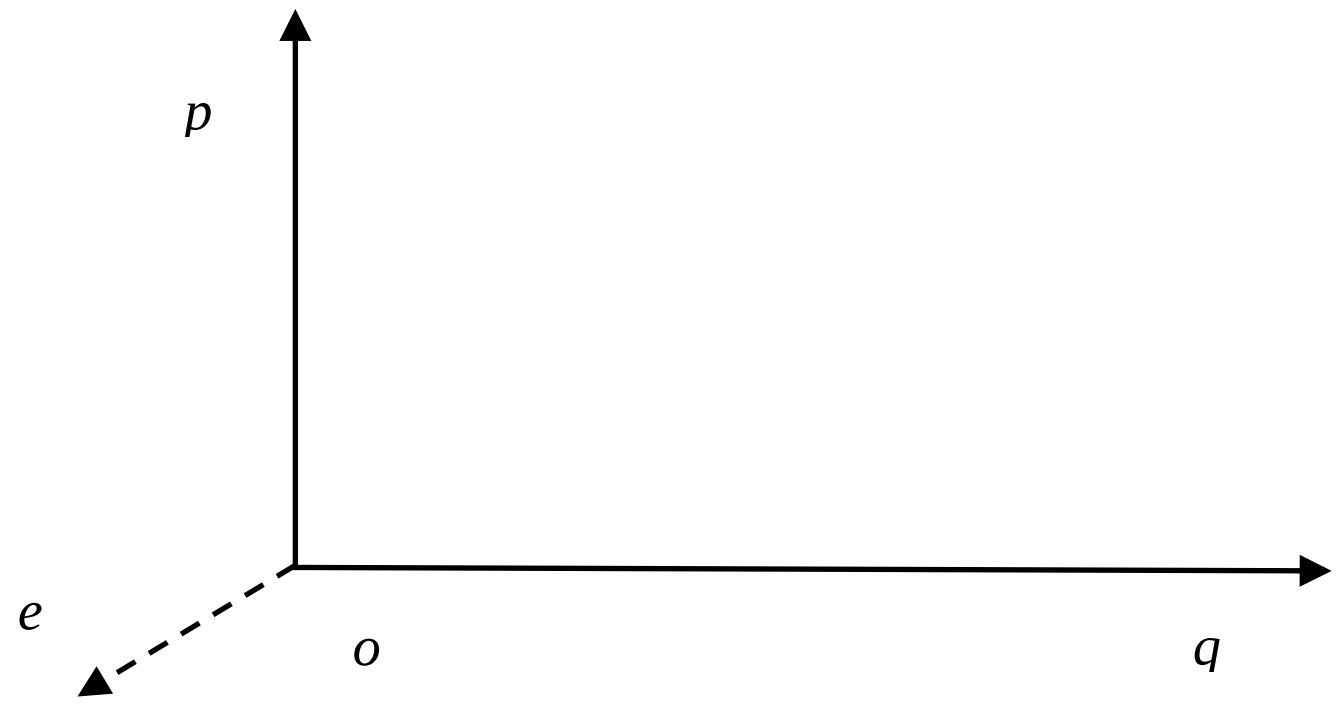

Fig. 4. Skinner-Shi's coordinates.

Note: ($q$) denotes a generalized reinforcement or sensitive variable; ($p$) is operant probability or frequency (likelihood or rate of a behavioral response); ($e$) represents a set of steerable multilevel eigenstates; ($o$) marks the origin.

Empirical analysis using this framework reveals that adaptive entangled game modes account for 82~94% (89% overall) of observed trading behaviors, a stark contrast to the predictions of neoclassical models based on independent rational agents. Additionally, adaptive dynamics to news, events, and environmental changes—characterized by dual equilibria and abrupt reference-point shifts—appear in 2-12% of cases, while purely independent behaviors are rare (<5%). These findings provide empirical support for the LCA hypothesis of entangled nerve fibers in the brain, since observable trading patterns reflect internal brain's decision-making and underlying mechanisms.

These results will enrich FMs by simulating brain mechanisms [52-54] and highlight the necessity of incorporating adaptive entangled game modules into AGI architectures, leading to the development of human-like processing units (HPUs) in the semiconductor and AI industries [55]. Such systems promise to be more compact, efficient, and robust than conventional ANN-based AI, especially for embodied intelligence and robotics.

Our main contributions are follows: (1) proposing a generalized probability-wave framework for modeling natural intelligence behaviors; (2) developing testable eigenmodels examining the LCS hypothesis through collective trading behaviors; (3) elucidating intelligence mechanisms through eigenvalues in eigenfunctions; (4) specifying necessary conditions for AGI architectures; and (5) theoretical guidance for semiconductor industry through simulating adaptive, entangled-brain mechanisms in the future AGI architectures.

We conclude with a discussion of the limitations and future directions for integrating both ANN-based AI and brain-inspired adaptive entangled models in probability-wave-based AGI systems.

## 2. Methods and Data

In this section, we introduce novel constructs for modeling human behaviors in intelligent decision-making: crowd operant momentum, operant force, and operant energy. We present a mutually influential operant energy equation and propose a probability-wave-based simulation theory for AGI through a generalized behavioral intelligence (GBI) probability-wave equation. This equation characterizes both independent and mutually influential, adaptive entangled game models through eigenvalues in associated wave functions, thereby revealing fundamental mechanisms of collective actions in complex adaptive systems.

According to an operant conditioning principle in behavioral psychology, observable collective actions reflect the internal brain's intelligent decision-making and mechanism. Thus, we can examine the LCA hypothesis through collective trading behaviors using Chinese stock market data.

### Data Description

To test collective trading behaviors and examine the LCA hypothesis of entangled nerve fibers in the brain, we utilize tick-by-tick trading data from Chinese stock markets for the years 2003, 2007-2009, 2019, and 2026. This real-world dataset provides a unique context for observing and simulating human intelligence operant behaviors in open dynamic environments (see Supplementary Material Information 1-4).

### Intelligent Decision-Making

Intelligent agents make decisions using a combination of heuristic rule construction, model learning or mimicry, and optimization strategies. At each decision point, these agents engage in inseparable, opposing, mutually influential, and adaptively entangled games—exemplified by collective action scenarios such as stock trading, where traders seek to buy low and sell high.

Extending from Skinner's operant conditioning [30, 31] and Shi's trading volume-price probability wave differential equation for Chinese stock markets [15], we generalize a generalized natural intelligence probability-wave framework. Here, agents respond to information and events with expectations, receive reinforcement or punishment via feedback, and adapt by adjusting their operant frequency or probability. Agents continuously adapt to their environment and persist through evolutionary mechanisms. The three-term contingency framework applies to collective traders in stock markets (see Fig. 3).

### Foundational Assumptions and Mathematical Framework

To construct a generalized behavioral intelligence (GBI) probability-wave equation in decision-making, we introduce the following foundational assumptions with corresponding mathematical expressions:

**Assumption Ⅰ:** Operant momentum ($Q$) at a reinforcement variable point ($q$) is defined as the cumulative operant quantity ($m$) at ($q$) over a time interval ($t$):

$$Q \equiv \frac{\partial S(q,t)}{\partial q} = \frac{m}{t} = m_t; \qquad (1)$$

**Assumption Ⅱ:** Operant force ($F_m$) is the rate of change of operant momentum ($Q$) with respect to time ($t$):

$$F_m \equiv \frac{Q(q,t)}{t} = \frac{m_t}{t} = \frac{m}{t^2} = m_{tt}; \tag{2}$$

**Assumption Ⅲ:** Operant energy ($E$) is the product of operant force ($F_m$) and the reinforcement variable ($q$):

$$E(q,t) = F_m * q = qm_{tt} = q\frac{m}{t^2} = q\frac{m_t}{t}. \tag{3}$$

Here, ($S(q, t)$) denotes collective operant action; ($q$) is the reinforcement or sensitive variable; ($m$) is the cumulative operant quantity; ($Q$) is the operant momentum; ($F_m$) is the operant momentum (dispersive) force; and ($E(q, t)$) is the operant energy. The time interval ($t$) is chosen for analytical convenience. These constructs formalize and quantify collective operant behavior within Skinner-Shi's coordinate system (see Fig. 4), serving as phenomenological tools to express interaction intensity, reference-point effects, and distributional structure in collective adaptive intelligence. They are not literal physical quantities but operational analogs for behavioral modeling.

**Interdependent Causation in Complex Adaptive Systems (CASs)**

Generalized behavioral intelligence systems, as complex adaptive systems, consist of many mutually influential agents who adapt through reinforcement learning and feedback loops. Interdependent causation develops between inputs and outputs via these loops. To capture this, we establish a mathematical relationship among operant energy, mutually influential energy, and reference-point potential.

The mutually influential energy equation is:

$$-E(q,t) + q\frac{m_t^2}{M} + U(q - q_0) = 0. \tag{4}$$

Here, ($E(q, t)$) denotes operant energy, ($PE(q, t)$) or ($q(m_t^2/M)$) represents mutually influential energy (originating from the Ising model [56]); and ($U(q\text{-}q_0)$) is the operant potential, with ($q_0$) as the reference point (see Appendix 1).

We further define a "linear" potential and the corresponding reversal force:

$$U(q) = A_{tt}(q - q_0), \tag{5}$$

and

$$F_r = -\frac{d}{dq}U(q) = -A_{tt}. \tag{6}$$

By substituting (3) and (5) into (4) and differentiating with respect to ($q$):

$$-m_{tt} + \frac{m}{M}m_{tt} + A_{tt} = 0, \tag{7}$$

where ($F_m$=$m_{tt}$) is the dispersive force, ($Fi$=$(m/M)m_{tt}$) is the mutually influential force, and ($F_r$=$-A_{tt}$) is the reversal force (negative sign indicates direction toward the reference point ($q_0$)).

**Generalized Behavioral Intelligence (GBI) Probability Wave Equation**

Drawing from Eq. (4), and applying mathematical methods analogous to Schrödinger's approach, we derive a time-independent GBI probability wave equation (see Appendix 2) [15-18]:

$$\frac{B^2}{M}\left(q\frac{d^2\psi}{dq^2} + \frac{d\psi}{dq}\right) + [E - U(q - q_0)]\psi = 0, \tag{8}$$

with normalization:

$$\int |\psi(p)|^2\, dp = 1 \text{ or } \sum_i |\psi_i(p_i)|^2 = 1 \tag{9}$$

where ($|\psi(q)|^2$) is the GBI probability wave function. This forms the core of the GBI probability wave framework.

## 3. Results and Empirical Tests

We will solve the GBI probability wave equation and examine the Liu-Chen-Ao hypothesis of entangled nerve fibers in the brain using tick-by-tick high-frequency data in Chinese stock markets in this section.

### Solving the GBI Probability Wave Equation

To solve (Eq. 8), we substitute the "linear" potential (Eq. 5), adopt natural units, and apply boundary conditions (Eqs. 10 and 11):

$$q\frac{d^2\psi}{dq^2}+\frac{d\psi}{dq}+[E-A_{tt}(q-q_0)]\psi=0, \qquad (10)$$

and

$$\begin{cases}\psi(+\infty) & \to 0\\ \psi(q_0) & <\infty\, limited.\\ \psi(-\infty) & \to 0\end{cases} \qquad (11)$$

These define GBI systems for empirical evaluation and allow for deriving both independent and mutually influential, adaptive entangled game models.

### Independent Operant Models

When operant energy ($E$) is constant, a family of multi-order explicit wave functions (eigenfunctions) describes independently behaving crowds (see Eqs. 12-18 and Fig. 5).

$$\psi_{n,i}(q_i)=C_l e^{-\sqrt{A_{tt,n,l}}|q_i-q_0|}\cdot F\left(-n,1,2\sqrt{A_{tt,n,l}}|q_i-q_0|\right), \qquad (n,l,i=0,1,2\cdots) \qquad (12)$$

or

$$\left|\psi_{n,i}(q_i)\right|^2=C_l e^{-2\sqrt{A_{tt,n,l}}|q_i-q_0|}\cdot\left|F\left(-n,1,2\sqrt{A_{tt,n,l}}|q_i-q_0|\right)\right|^2, \qquad (n,l,i=0,1,2\cdots) \qquad (13)$$

subject to

$$\sqrt{A_{tt,n}}=\frac{E_n}{1+2n}=\text{constant}>0, \qquad (n=0,1,2\cdots) \qquad (14)$$

or

$$E_n=(1+2n)\sqrt{A_{tt,n}}=2\left(n+\frac{1}{2}\right)\omega_n=const.>0 \qquad (n=0,1,2\cdots) \qquad (15)$$

and

$$F(\alpha,\gamma,\xi)=1+\frac{\alpha}{\gamma}\xi+\frac{\alpha(\alpha+1)}{2!\gamma(\gamma+1)}\xi^2+\frac{\alpha(\alpha+1)(\alpha+2)}{3!\gamma(\gamma+1)(\gamma+2)}\xi^3+\cdots=\sum_{k=0}^{\infty}\frac{(\alpha)_k}{k!(\gamma)_k}\xi^k, \qquad (16)$$

$$(\alpha)_k=\alpha(\alpha+1)\cdots(\alpha+k-1), \qquad (17)$$

$$(\gamma)_k=\gamma(\gamma+1)\cdots(\gamma+k-1). \qquad (18)$$

Eq. (13) represents independent operant probability models, while Eq. (14) or (15) capture the underlying quantized energy eigenstates.

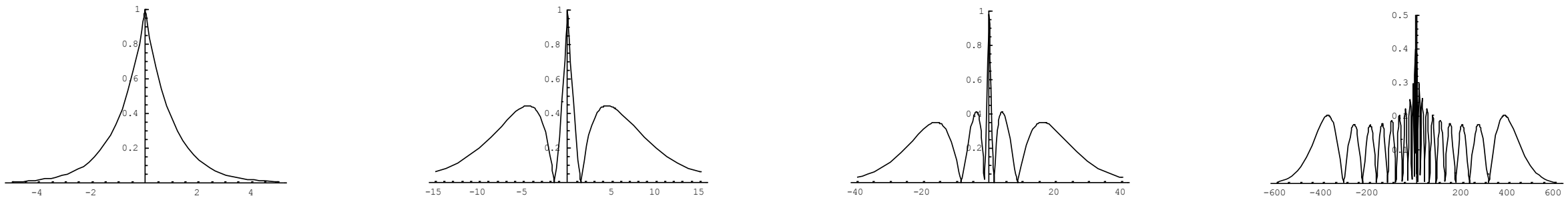

(a) (b) (c) (d)

Fig. 5 Independent intelligence operant models (n=0, 1, 2, 10)

Note: Horizontal axis—reinforcement variable; vertical axis—intelligence operant frequency/probability over time; origin—decision-making reference point.

### Mutually Influential, Adaptive Entangled Game Models

When operant energy ($E=pv_{tt}$) is separable, mutually influential, adaptive entangled game models arise (see Eqs. 19-21, Fig. 6):

$$\psi_{n,i}(q_i) = C_n J_{0,i}[\omega_n(q_i - q_0)], \qquad (n = 0,1 \cdots), (i = 1,2 \ldots) \tag{19}$$

or

$$\left|\psi_{n,i}(q_i)\right|^2 = C_n \left|J_{0,i}[\omega_n(q_i - q_0)]\right|^2, \qquad (n = 0,1 \cdots), (i = 1,2 \ldots) \tag{20}$$

Subject to:

$$\omega_n^2 = \frac{m_{t,n,i}^2}{M} = \frac{m_{n,i}}{M} m_{tt,n,i} = m_{tt,n,i} - A_{tt,n,i} = F_m - F_r = const. \qquad (n = 0,1,\cdots), (i = 1,2 \ldots) \tag{21}$$

Eq. (20) simulates mutually influential, adaptive entangled game behaviors, where ($J_{0,i}[\omega_n(q_i - q_0)]$) is a set of zero-order Bessel eigenfunctions and ($C_n$) is a normalization constant. Eqs. (19)-(21) define the Bessel-Shi models for such game modes.

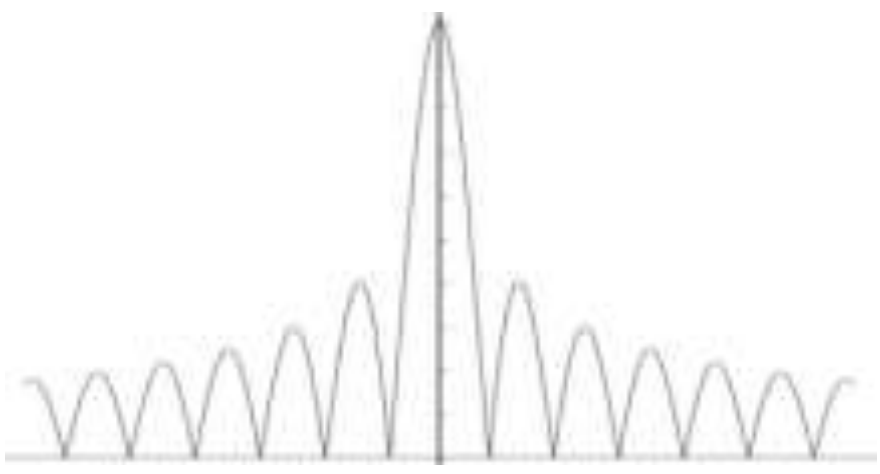

(a)

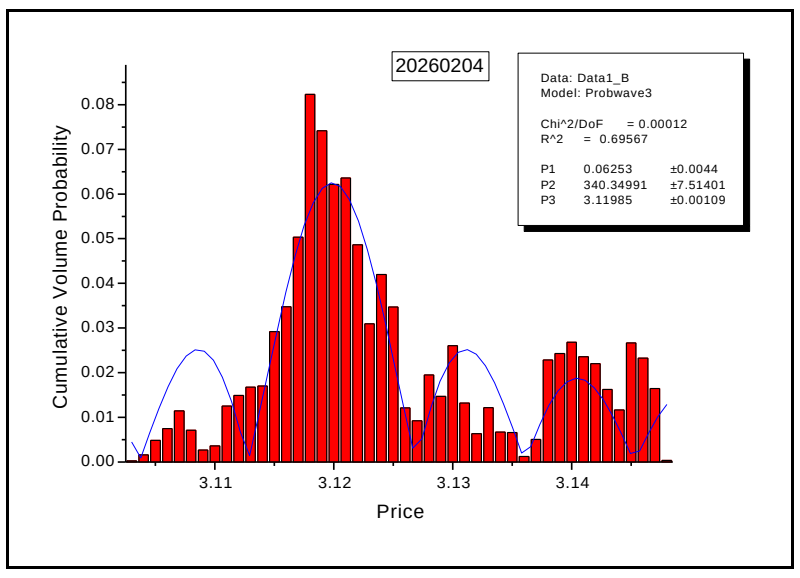

(b)

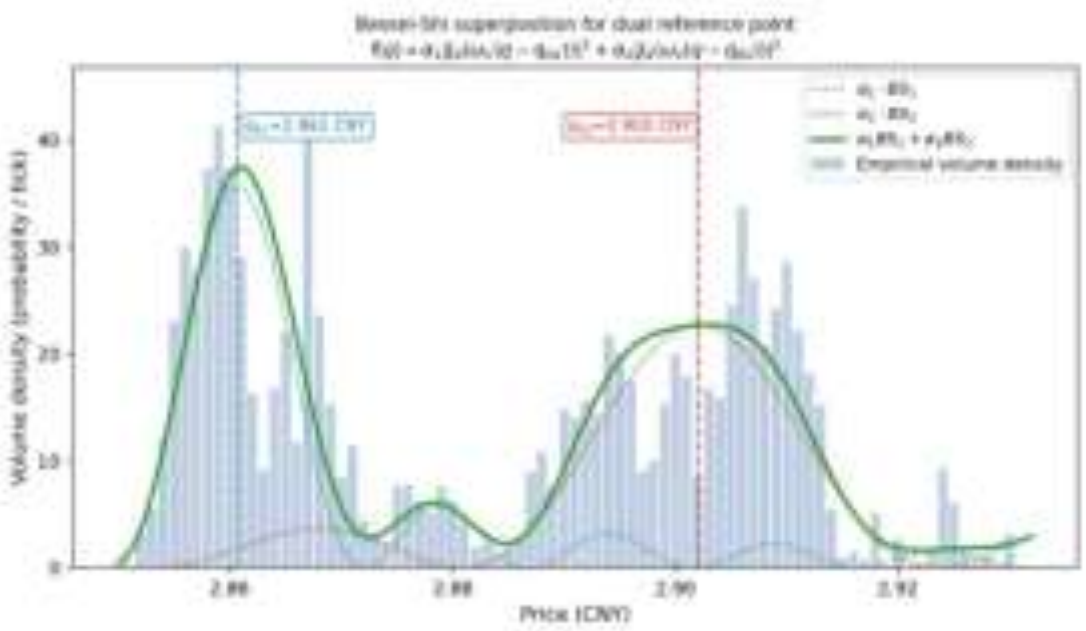

(c)

Fig. 6. Adaptive entangled Bessel-Shi models.

Note: (a) generalized reinforcement variable (horizontal axis), intelligence operant frequency/probability (vertical axis), reference point at origin; (b) simulation of adaptive entangled game patterns in stock markets (Feb 4, 2026)—(*P1*): maximal probability at reference point (*P3=RMB ￥3.12*), (*P2*): interaction-coherent eigenfrequency (*$\omega_n$=340.35>0*), significance at 95% confidence, ($R^2$=0.70>$R^2_{crit}$=0.08); (c) adaptive dynamics to news, events, and environments—characterized by dual reference points (RMB￥2.86 and ￥2.90), including a reference-point jump (Mar 23, 2026). The screening threshold is the *F*-test critical value of Eq. (22), evaluated at ($\alpha$) = 0.05, ($k$) = 3 fitted coefficients ($C_n$, $\omega_n$, $q_0$) and ($n$) = 74 price levels, the mean session length in the 2007-2009 record; panel (c) is a six-coefficient fit and is screened against its own, higher critical value ($R^2_{crit}$ = 0.147). See "Fitted Parameters, Degrees of Freedom, and the Significance Threshold".

**Empirical Tests**

Applied an operant conditioning principle in behavioral psychology, we tested indirectly an entangled brain hypothesis through collective trading behaviors in Chinese stock markets.

We analyzed four intraday tick-by-tick trading datasets from the Chinese stock markets (2003, 2007-2009, 2019, 2026; see Supplementary Material Information 1-4), using pattern analysis, time-series model-agnostic analysis, and data-driven analyses. Baseline model comparison involved probability-wave-based Bessel-Shi models, unimodal statistical baselines (Normal and Lognormal), and a two-component Gaussian mixture (GMM2). Pattern analysis involved fitting intraday cumulative trading volume across price ranges to our probability-wave-based models, with significance evaluated using $R$-squared against the $F$-test critical value at the 95% confidence level. In addition, we conduct unified comparisons using metrics such as AIC, BIC and cross-validated held-out (out-of-sample) likelihood.

**Pattern Analysis**

Empirical testing proceeded in three rounds using Origin 7.0. First round: Applied adaptive entangled Bessel-Shi models (Eqs. 20-21) to all datasets (see Fig. 6b). Second round: For non-significant cases, captured adaptive behavior with dual reference points via a superposition of two Bessel-Shi wave functions (see Fig. 6c). Third round: Remaining cases were evaluated using independent intelligence operant models (Eqs. 13-15).

A “multi-center structure” refers to distributions with more than one concentration center in statistical analysis (see Fig. 6b). “Dual reference points” denotes two distinct equilibrium points in a probability wave context (see Fig. 6c). Table 1 is a pattern analysis report in empirical tests.

**Table 1: Pattern Analysis Reports**

| | 2003 | 2007-2009 | 2019 | 2026 | Overall Results |
|---|---|---|---|---|---|
| **Total No. in Tests** | 618 | 495 | 34 | 36 | 1183 |
| **1st Round Tests** | | | | | |
| Number of pass in test | 583 | 408 | 32 | 31 | 1054 |
| Number of no pass | 35 | 87 | 2 | 5 | 129 |
| Pass ratio | **94.34%** | **82.42%** | **94.12%** | **86.11%** | **89.10%** |
| **2nd Round Tests** | | | | | |
| Number of pass in test | 34 | 59 | 1 | 4 | 98 |
| Number of no pass | 1 | 28 | 1 | 1 | 30 |
| Pass ratio | **5.50%** | **11.92%** | **2.94%** | **11.11%** | **8.28%** |
| **3rd Round Tests** | | | | | |
| Number of pass in test | 1 | 23 | 1 | 0 | 25 |
| Number of no pass | 0 | 5 | 0 | 1 | 6 |
| Pass ratio | **0.16%** | **4.65%** | **2.94%** | **0.00%** | **2.11%** |

Note: (1) Analysis covers 4 datasets from Supplementary Data 1-4; (2) 82-94 % (89% overall) of decision-making patterns matched adaptive entangled game modes; (3) Pass/fail is defined against the fixed threshold $(R^2) \geq R^2_{crit}$ (Eq. 22 at ($k$) = 3, ($n$) = 74). The three rounds fit models with 3, 6 and 3 free coefficients, respectively, so the three pass ratios form a sequential screen and not a ranking; degrees-of-freedom-aware re-screening and a unified AIC/BIC/cross-validation comparison are reported in Table 3.

**Table 2: Prevalence of Statistically Supported Multi-Center Structure with Bootstrap Uncertainty**

| **Year** | Multi-Center Ratio | **95% CI** | **Shuffle Ratio** | **95% CI (shuffle)** |
|---|---|---|---|---|
| **2007** | 0.1027 | [0.00595, 0.1459] | 0.027 | [0.0054, 0.0541] |
| **2008** | 0.0816 | [0.0490, 0.1184] | 0.0449 | [0.0204, 0.0694] |
| **2009** | 0.0244 | [0.0049, 0.0488] | 0.0098 | [0.0000, 0.0244] |

Note: Prevalence analyzed by period for Huaxia SSE50 ETF (510050; 2007-2009), using bootstrap 95% confidence intervals and shuffle-control comparisons. These estimates are descriptive and should not be conflated with the proportion of sessions with dual reference points (see Fig. 6).

### Time-Series Model-Agnostic Analysis

To address concerns about reliance on a single metric (e.g., R-squared) or functional form (e.g., the Bessel–Shi model), we conducted model-agnostic analyses of time-series tick-by-tick trades for Huaxia SSE50 ETF (510050; 2007-2009; 635 sessions). We compared Bessel-Shi models to traditional unimodal baselines (normal and lognormal) and a two-component Gaussian mixture model (GMM2), using weighted maximum likelihood. Results show that empirical distributions exhibit much stronger probability-wave coherent structure than randomized controls (see Fig. 7, Table 2).

These tests were supplemented with information-criterion baseline comparisons, bootstrap uncertainty quantification, and shuffle falsification controls. Additional analyses examined statistical rigor, underlying mechanisms, predictive power, independence, and external validity, with all modeling done in Python.

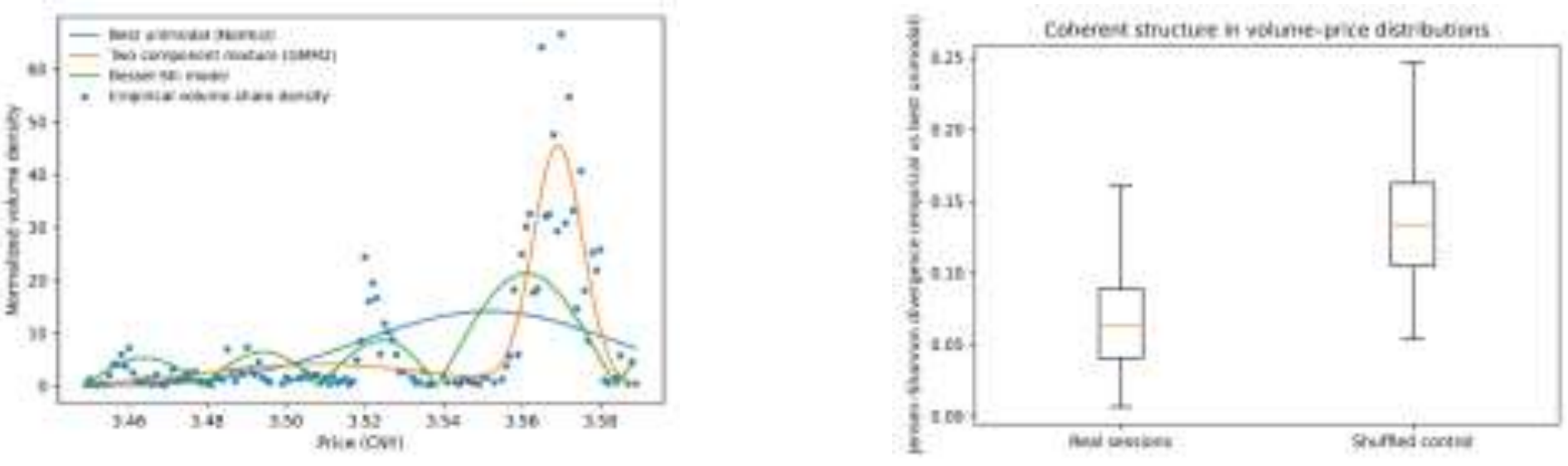

(a) (b)

Fig. 7. Comparison of the adaptive entangled game model (Bessel-Shi) with traditional baselines and GMM2.

Note: (a) “Multi-center” structure revealed by all models; (b) Jensen-Shannon Divergence (JSD) in real sessions vs. shuffled controls quantifies departure from unimodal shape, with high JSD indicating stronger volume-price coupling.

### Data-Driven Analysis

To further assess rigor, mechanisms, predictive power, independence, and external validity, we performed five additional data-driven analyses on the full tick-by-tick record of the Huaxia SSE50 ETF (510050; 2007–2009), all using pre-registered, falsification-style controls (cross-validation or volume shuffling) from cumulative-volume snapshots (see Figs 8 and 9; Table3).

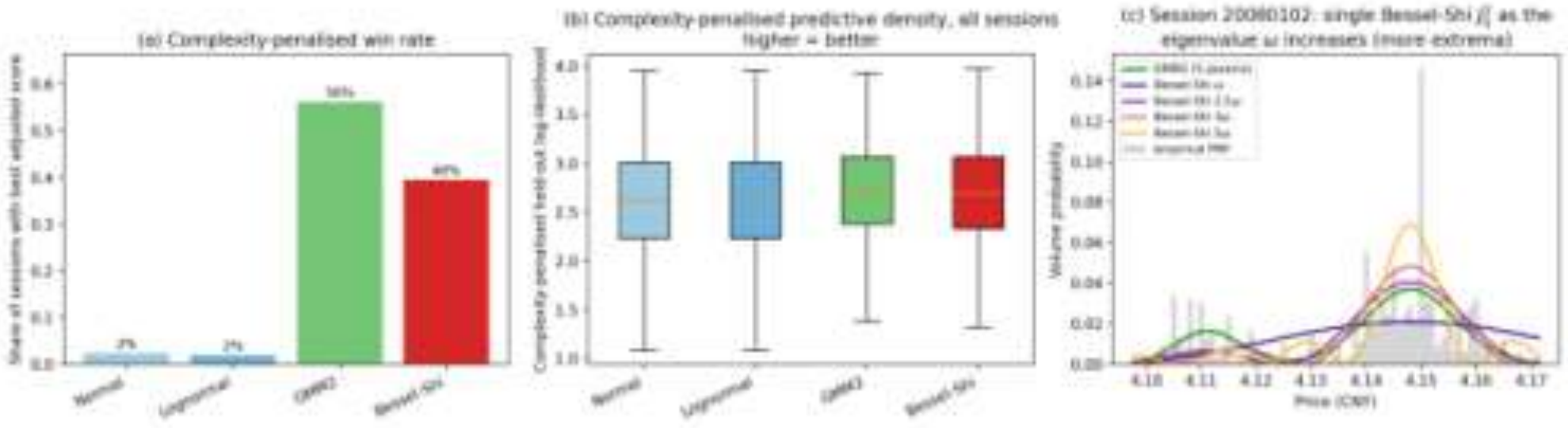

Figs. 8. Out-of-sample model selection

Note: (a) Share of sessions for which each model achieves best held-out log-likelihood; (b) distribution of held-out log-likelihood by models; (c) Bessel-Shi model reproduces multi-center structure parsimoniously (2 explicit parameters versus 5 unknown parameters for GMM2, or up to 11 when BIC selects components).

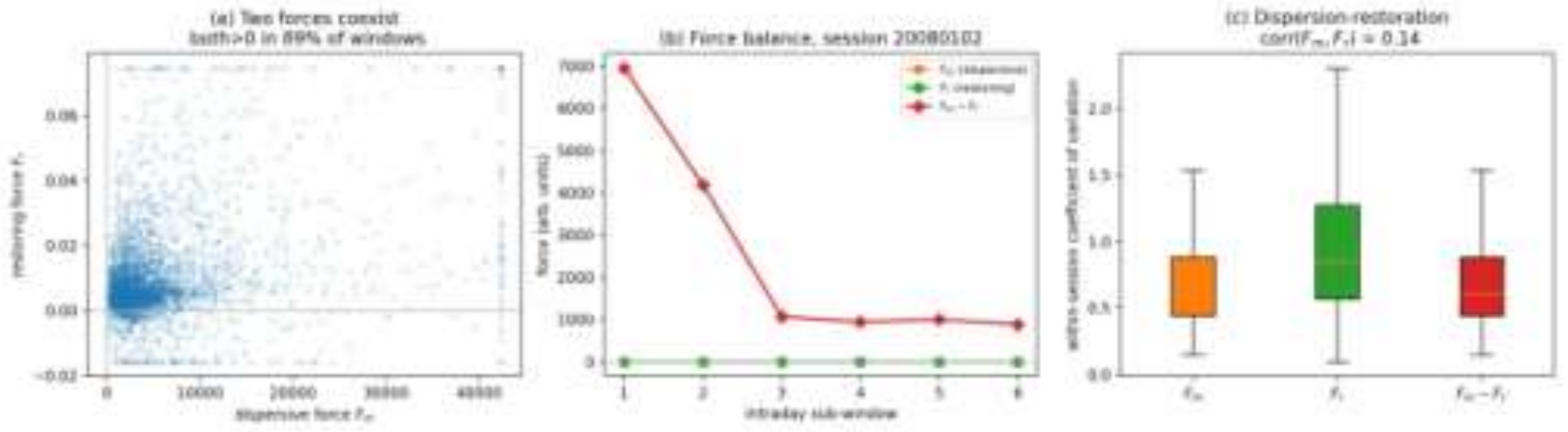

Fig. 9. Dispersion-restoration balance.

Note: (a) Joint distribution of the dispersive ($F_m$) and restoring ($F_r$) forces; (b) example session; (c) within-session coefficient of variation of ($F_m$), ($F_r$), and ($\omega_n$=$F_m$−$F_r$), representing underlying mechanisms in interaction-coherent entanglement.

These analyses (1) replace the single-metric ($R^2$) rule with complexity-penalized, out-of-sample model selection; (2) provide direct measurements of dispersive and restoring operant forces; (3) directly test the reference-point mechanism; (4) reject independent-agent null hypotheses; and (5) demonstrate regime dependence of mutual influence intensity. Collectively, these results robustly support the empirical basis for adaptive entangled game behaviors, which are discussed in the following section.

**Unified Model Comparison**

The three rounds of pattern analysis, the model-agnostic tests and the data-driven analyses use different fitting protocols and different acceptance rules, so we finally place all candidates on a single footing. For each of the 635 Huaxia SSE50 ETF sessions (2007-2009) we fit the truncated Normal, the truncated Lognormal, GMM2, the single Bessel-Shi wave and the dual Bessel-Shi superposition as normalized densities on a common support, and score them by (i) the volume-weighted log-likelihood, (ii) AIC and BIC with the effective sample size taken as the number of distinct price levels, and (iii) three-fold cross-validated held-out log-likelihood, with folds taken over price levels so that no level contributes to both fitting and scoring. Table 3 collects the main comparison results across all three analyses.

**Table 3: Summary of the Main Comparison Results**

| Category | Comparison/Measure | Result/Key Value(s) |
|---|---|---|
| **Pattern Analysis** | Single Bessel-Shi (3 coeff.) | 89.1% overall (82.4% in 2007–09) |
| | Refitted by max likelihood | 69.8% (fixed), 72.0% (d.o.f.-aware) |
| | Dual-wave (6 coeff.) | 84.7% (fixed), 73.2% (d.o.f.-aware) |
| **Model-Agnostic Evaluation** | Bessel-Shi vs. unimodal baselines | 92.9% sessions with higher LLH |
| | GMM/KDE vs. Bessel-Shi | GMM: 86.5% highest held-out LLH |
| **Unified Model Comparison** | Model parameters (curve/density) | Bessel-Shi: 3/2 or 6/5; GMM2: 6/5 |
| | BIC win rate (1-wave) | 63.5% |
| | CV win rate (1-wave) | 39.5% |
| | AIC win rate (dual-wave) | 35.4% |
| **Mechanism Tests** | Both forces positive | 88.9% of windows |
| | Reference-point reversion ($\kappa$) | Positive in 96.5% of sessions |
| | Trade-sign ACF(1) real/shuffled | -0.430 / ~0 |
| | MII & volatility correlation | 0.372 |
| | Regime dependence (ANOVA) | F = 45.2, p = 4.4e-19 |
| **Directional Prediction** | Reference-point | 0.507 accuracy |
| | Momentum | 0.561 accuracy |

Note: Except for the first row, which reproduces Table 1 (2003, 2007-2009, 2019 and 2026, fitted by least squares in Origin 7.0), every entry is computed on the full tick record of the Huaxia SSE50 ETF (510050), 2007-2009, 635 sessions. The

maximum-likelihood refit of the first-round model optimizes predictive density rather than squared error, so its pass rate (69.8%) is a lower bound on the least-squares figure (82.4%) rather than a reproduction of it. Effective sample size for AIC and BIC is the number of distinct intraday price levels in the session (mean 74, median 67); cross-validation is three-fold over price levels, so no level contributes to both fitting and scoring. Confidence intervals are nonparametric bootstrap percentiles over sessions. A negative ΔBIC would favor the dual wave; the positive median reported here means the second-round superposition does not pay for its three extra degrees of freedom on the typical session. The mechanism tests of Figs. 9-12 are reported for completeness; the directional-accuracy entry is a negative result and is discussed under Limitations.

## 4. Discussions

Simulation theories of the human brain—most notably artificial neural networks (ANNs) [44, 45]—have profoundly advanced artificial intelligence, underpinning the development of large language models (LLMs) and reinforcement learning-based LLMs (RL-LLMs) from OpenAI and DeepSeek. However, ANN-based AI relies on processing vast amounts of information through network structures comprising trillions of opaque parameters, resulting in high computational and energy costs, as well as limited interpretability. In response, AI scientists have introduced the concept of AGI and developed FMs [11, 12]. Yet we still lack a clear understanding of how they function, when they fail, and what they are truly capable of, due to their emergent properties.

Understanding the mechanism of the human brain remains a central focus in neuroscience, and simulating these processes represents a major frontier in AGI research. Liu, Chen and Ao identified quantum entanglement as a potential explanation for how millions of brain cells synchronize their activity to enable coherent function [27, 28]. Deco, Perl, and Kringelbach utilized Schrödinger's wave equation to model the nonlocality inherent in brain activity [29]. In parallel, Shi et al. discovered a nonlocal many-body wave equation in complex quantum systems [18], directly characterizing nonlocal behaviors within complex adaptive systems. Building on this sequence of research, our paper proposes a nonlocal probability-wave framework for understanding generalized behavioral intelligence, mainly focusing on adaptive entangled behaviors in the brain and advancing FMs for AGI architectures.

While this nonlocal, entangled brain mechanism presents intriguing possibilities, direct experimental validation remains challenging. To address this, we propose a nonlocal probability-wave theory of generalized behavioral intelligence with interpretable mechanisms, applying it to indirectly test the LCA hypothesis through collective trading behaviors using tick-by-tick high-frequency data from Chinese stock markets. Our empirical analysis of Chinese intraday stock market data reveals that 82-94 % (overall 89%) of decision-making patterns aligns with mutually influential, adaptive entangled modes. An additional 2–12% of behaviors display adaption to news, events, and environments, characterized by dual reference points and abrupt state shifts, while fewer than 5% correspond to independent modes. Notably, across the 2007-2009 market cycle—including periods of bubble growth, collapse, and reversal, coinciding with the 2008 U.S. subprime crisis—market participants predominantly exhibited adaptive, entangled, mutually influential behaviors. These results challenge the neoclassical finance assumption of independent, rational agents and highlight the prevalence of interaction-driven, nonlocal, adaptive entangled game dynamics. Our findings empirically support the LCA hypothesis (see Tables 1-3).

Importantly, human brain entangled states belong to nonlocal interaction-coherent entangled states [18] rather than energy-quantum entangled states. Whereas the latter occur in extreme environments (e.g., near absolute zero), the nonlocal interaction-coherent entangled states are characterized by eigenfrequency conservation and can be observed in both stock markets and the brain.

The nonlocal generalized behavioral intelligence (GBI) probability wave differential equation, extended from the trading volume-price probability wave equation developed in 2006, contrasts with conventional probability theory as used in random process modeling—such as Deng's 2026 Fields work [57] and Boltzmann's kinetic equation [58]. Whereas conventional probability theory describes randomness using opaque parameters, the probability wave approach models uncertainty through interpretable eigenvalues of wave eigenfunctions, enabling steerable, interpretable system states.

ANN-based AI models can be traced back to the Ising model, which has a wide range of applications—from fixed interactions, such as Wilson's renormalization and Parisi's studies of disorder and fluctuations in physical systems, to individually adjustable interactions [59], including the Hebbian-inspired rule [44] and constructive divergence rule [60]. In contrast, generalized probability-wave-based simulations rely on a mutually influential energy equation (4), where the sum of interaction energy and a linear potential defines the Hamiltonian. This study draws inspiration from the LCA hypothesis in biophysics, the trading volume-price probability wave equation in behavioral econophysics, and the nonlocal many-body wave equation in quantum systems, all unified mathematically under the theory of complex adaptive systems.

Based on empirical results, we encourage the exploration of human-like processing units (HPUs) that incorporate adaptive entangled functions within the semiconductor industry, with potential applications in autonomous driving, robotics, and embodied intelligence [10, 55]. For AGI systems to accurately simulate diverse human behaviors, they must integrate GBI modules—especially adaptive entangled game modules for collective actions. The eigenvalues of wave eigenfunctions reveal these underlying mechanisms, suggesting that behavioral intelligence modules will be crucial for realizing multimodal AGI systems (Fig. 10).

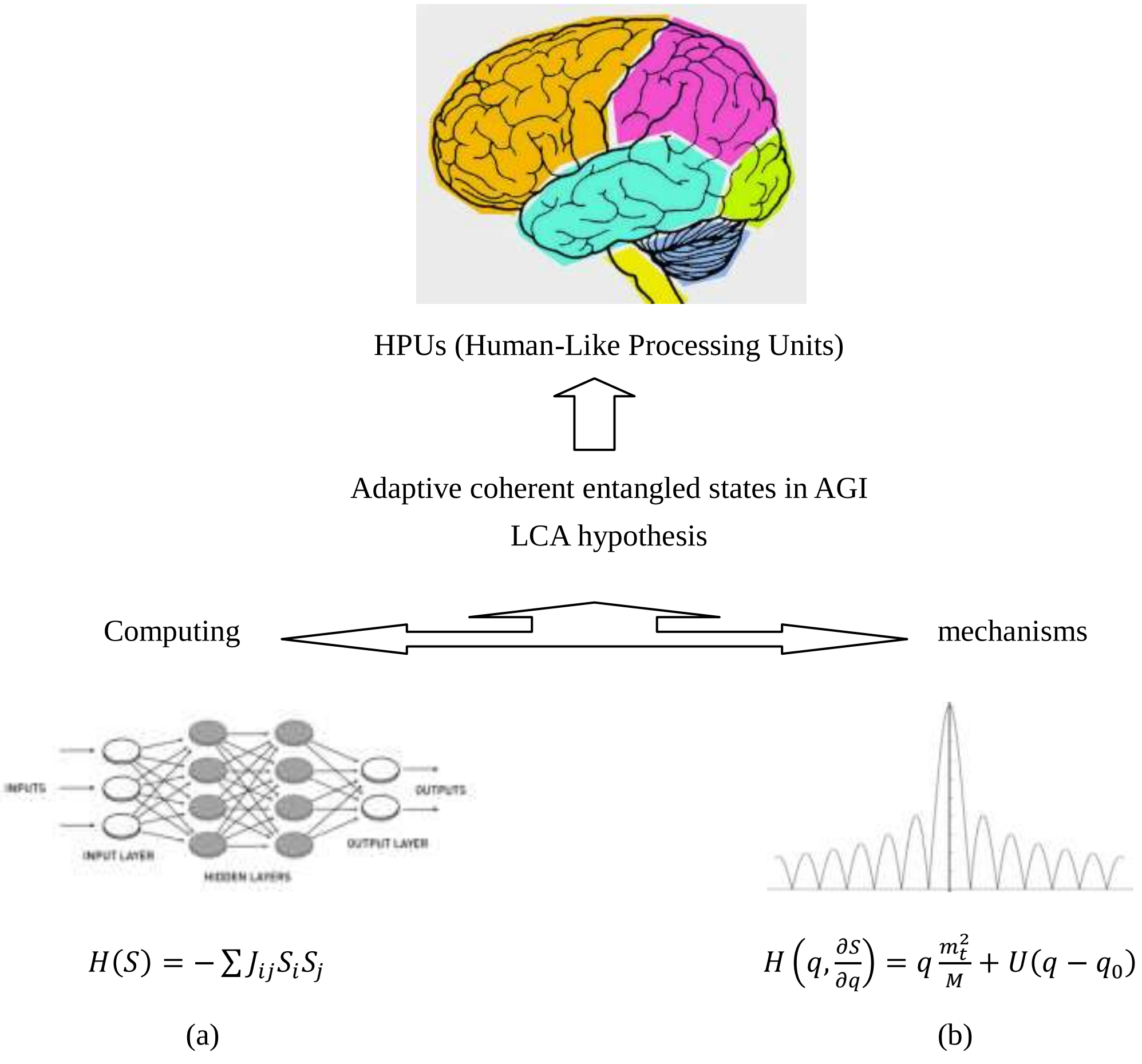

Fig. 10. Simulations of human-like brain in machine learning

Note: (a) The left-hand side shows Hamiltonian $H(S) = -\sum J_{ij} S_i S_j$ in the Ising model, where (*Jij*) quantifies the interaction strength between nearest neighbors *i* and *j* and (*Si*) $\in$ {−1, 1} corresponds to the spin orientation at site i. (b) The right-hand side presents Hamiltonian $H(q, \partial S/\partial q) = qm_t^2/M + U(q - q_0)$, which is the sum of interaction energy and linear potential.

We further compare our probability wave-based models to conventional statistical baselines, including unimodal distributions (Normal and Lognormal) and a two-component Gaussian mixture model (GMM2). The principal advantage of the Bessel-Shi model lies in its mechanism-level interpretability, rather than statistical fit alone. While conventional models can capture localized data features, the Bessel-Shi family is intrinsically linked to the eigenfunction structure of mutually influential, adaptive entangled games. This family demonstrates higher out-of-sample win rates, improved predictive density (mean-held-out log-likelihood), deeper mechanistic insights, and a closer theoretical correspondence between observed market structures and the proposed intelligence probability-wave paradigm (Fig. 8).

Once model complexity is accounted for, the Bessel-Shi family is preferred over both classes of baseline on a session-by-session basis, though it does not always outperform more flexible estimators. Across 635 sessions, the single wave model achieves a lower BIC than GMM2 in 83.6% of sessions and outperforms the best unimodal baseline in 87.1%. It also attains a higher cross-validated likelihood than GMM2 in 63.1% of sessions. When the candidate set is expanded to include a kernel density estimate and mixtures with up to eleven free parameters, the mixture models achieve the highest raw held-out likelihood in 86.5% of sessions (Fig. 8, Table 3). Thus, the case for the Bessel-Shi model rests on its ability to provide a comparable description with just two mechanism-driven parameters, as well as independent force-balance, reference-point, independence, and regime evidence, rather than on outperforming more flexible curves in goodness-of-fit contest (see Table 3).

From empirical tests, we conclude:

- The Bessel-Shi (probability-wave) model consistently outperforms unimodal baselines in both pattern fit ($R^2$ pass rates) and out-of-sample log-likelihood across most sessions, demonstrating strong empirical validity.
- Flexible mixture models (such as GMM2) often achieve the highest raw held-out likelihood, but at cost of increased model complexity and reduced mechanistic interpretability compared to the Bessel-Shi models.
- Model selection criteria (AIC, BIC and cross-validation) generally favor the single-wave Bessel-Shi model over both more complex dual-wave and mixture models, indicating a good balance between fit and parsimony.

Overall, Table 3 shows that the adaptive probability-wave (Bessel-Shi) framework provides a robust, interpretable description of the observed data, outperforming traditional models by most criteria while retaining meaningful mechanistic insights.

Limitations:

First, while the adaptive entangled game module enhances our understanding of AGI mechanisms, the present study does not encompass the full breadth of behavioral intelligence. Second, direct application and comparative evaluation against existing algorithms in diverse real-world settings are reserved for future work. Third, empirical validation has so far been limited to Chinese stock market data; extending this work to other markets and behavioral domains is an important next step. Fourth, it remains an open question how the probability wave simulation theory of GBI can dramatically reduce the number of opaque parameters without sacrificing efficiency or function in practical AI and AGI systems. Next, it remains open to directly test the LCA hypothesis through neuron activity distributions in the brain against the Bessel-Shi model. Finally, the framework describes how collective decisions are organized, rather than forecasting their direction: the reference-point signal carries no out-of-sample directional predictability (accuracy 0.507 on a chronological 2007-2008/2009 split), consistent with the predictability boundary identified in Table 3.

## 5. Conclusions

ANN-based AI theory has laid the groundwork for today's powerful machine learning systems. However, its reliance on vast numbers of opaque parameters impedes our understanding of embodied intelligence and advanced robotics within AGI.

We propose a probability wave theory of GBI featuring interpretable mechanisms and apply it to test the LCA hypothesis through collective trading behaviors using tick-by-tick high-frequency data from Chinese stock markets. From the GBI probability wave equation, we derive both independent and mutually influential adaptive entangled game models. In adaptive entangled scenarios, the sum of the momentum (dispersive) force and the reversal force equals the interaction-coherent eigenfrequency, maintaining system stability across reinforcement variables. Understanding these mechanisms aids in preparing adaptive entangled states in machine learning to simulate the human brain and realize a multimodal AGI system.

Empirical analysis of Chinese intraday stock market data shows that 82-94% (89% overall) of trading decisions follow adaptive entangled game patterns—in sharp contrast to the independent rational agent model of neoclassical finance. Adaptive behaviors with dual equilibria and abrupt reference-point changes account for 2~12% of cases, while truly independent modes comprise less than 5%. These results challenge the neoclassical finance assumption of independent, rational agents, highlighting the dominance of interaction-driven, adaptive entangled game dynamics in stock markets and providing empirical support for the LCA hypothesis.

Integrating ANN-based AI with the brain mechanism simulations in machine learning will enrich FMs. It is essential for future AGI architectures. This integration offers theoretical guidance for developing AGI systems. Theoretical analysis suggests that HPUs built upon brain mechanisms could provide a promising pathway toward compact, efficient, and robust AGI systems, particularly in embodied intelligence and robotics.

Future challenges include: (i) directly testing the LCA hypothesis in neuroscience and examining interaction-coherent entanglement in nonlocal many-body systems; (ii) developing formal languages for communication between embodied intelligence and probability wave-based GBI; (iii) creating algorithms that bridge GBI and AGI; (iv) applying quantum information technology to generate long-life, adaptive entangled states for AGI; (v) inventing HPU manufacturing technologies for the semiconductor industry; and (vi) demonstrating how the probability wave simulation theory can advance AGI for embodied intelligence and robotics, etc.

## Acknowledgments

We are grateful for insightful discussions with Wu-Ming Liu, Ping Ao, Zefei Liu, Xinqiang Huang, Zhiyuan Liu, Guilu Long, Hao Wu, and others. We assume full responsibility for any omissions or errors in this work.

## Appendix 1: Mutually Influential Energy Equation

Starting from a general formulation applicable to any uncertain system, we make the following assumptions:

$$E(q,t) \equiv PE(q,t) + (1-P)E(q,t) = \frac{m}{M}E(q,t) + U(q-q_0), \tag{A1-1}$$

and

$$P = \frac{m}{M}. \tag{A1-2}$$

Here, ($q$) denotes a generalized reinforcement or sensitive variable, ($q_0$) is the zero point or center of the “linear” potential ($U(q)$); ($m$) is the cumulative operant quantity at ($q$) during time interval ($t$); ($M$) is the total cumulative operant quantity over ($t$) across the range of ($q$); ($P$) denotes the probability/frequency at ($q$); ($E(q,t)$) is the operant energy at ($q$); ($PE(q,t)$) is the cross energy or mutually influential energy over ($t$); and ($U(q)$) is the “linear” operant potential.

The cross energy term ($(m/M)E(q,t)$) in Eq. (A1-1) denotes mutually influential energy, a concept traceable to the Ising model [56].

By substituting Eqs. (3) and (5) into the right side of (A1-1), we obtain:

$$E(q,t) = PE(q,t) + U(q-q_0) = \frac{m}{M}\left(q\frac{m}{t^2}\right) + U(q-q_0) = \frac{1}{M}q\left(\frac{m}{t}\right)^2 + U(q-q_0). \tag{A1-3}$$

Eq. (A1-3) can be rearranged to yield Eq. (4). Both Eqs. (4) and (A1-3) apply to complex adaptive systems, such as GBI systems, capturing interdependent causation between inputs and outputs.

## Appendix 2: Generalized behavioral intelligence probability wave equation

Consider an unknown function ($\psi(q,t)$) expressed as [15-18]:

$$\psi(q,t) = Re^{iS(q,t)/B}, \tag{A2-1}$$

Here, ($q$) is a generalized reinforcement/sensitive variable, ($t$) is the time interval, ($R$) is the amplitude, ($S(q,t)$) is the operant action (Hamilton’s principal function, assumed twice differentiable), ($B$) is a dimensional constant ensuring the phase ($S(q,t)/B$) is dimensionless, ($i$) is the imaginary unit ($i^2 = -1$).

From Eq. (A2-1), we derive:

$$\frac{\partial S(q,t)}{\partial q} = -\frac{iB}{\psi(q,t)}\frac{\partial \psi(q,t)}{\partial q}. \tag{A2-2}$$

Assuming:

$$\frac{\partial S}{\partial t} + H\left(q, \frac{\partial S}{\partial q}\right) = 0, \quad \text{(A2-3)}$$

where ($H\left(q, \frac{\partial S}{\partial q}\right)$) is the Hamiltonian operator.

The Hamiltonian is given by:

$$H\left(q, \frac{\partial S}{\partial q}\right) = q\,\frac{m_t^2}{M} + U(q - q_0). \quad \text{(A2-4)}$$

Eq. (A2-3) is strictly valid for energy-conservative systems. Nonetheless, we adopt Eq. (A2-3) as an effective phenomenological extension for open GBI systems, with its validity judged empirically, for three reasons. First, Eq. (4) is derived from the general Eq. (A1-1) and remains applicable without strict energy conservation; thus, the operant-wave formulation is not restricted to conservative systems. Second, in certain limiting regimes (e.g., when (E) is nearly a constant), the conservative-form approximation is reasonable even for open systems. Most importantly, this assumption is falsifiable: if the GBI wave equation (as instantiated in eigenfunction/eigenvalue models) fails to reproduce observed crowd operant dynamics—including robustness and out-of-sample validation—then the assumption underlying Eq. (A2-3) should be rejected in favor of a more general open-system formulation.

By separating variables, Eq. (A2-3) becomes:

$$H\left(q, \frac{\partial S}{\partial q}\right) = -\frac{\partial S}{\partial t} = E, \quad \text{(A2-5)}$$

This leads to two equations:

$$-\frac{\partial S}{\partial t} = E, \quad \text{(A2-6)}$$

and

$$q\,\frac{m_t^2}{M} + U(q - q_0) = E. \quad \text{(A2-7)}$$

From Eq. (A2-6):

$$S(q, t) = S_1(q) - Et. \quad \text{(A2-8)}$$

A special solution can be written as:

$$S(q, t) = \alpha(q m_t) - Et + \beta \equiv q m_t - Et, \quad \text{(A2-9)}$$

where ($\alpha$) or ($\beta$) are arbitrary constants.

Thus, we define operant momentum ($Q$) and operant force ($F_m$) from Eq. (A2-9) as:

$$Q \equiv \frac{\partial S}{\partial q} = m_t, \quad \text{(A2-10)}$$

and

$$F_m \equiv \frac{Q}{t} = m_{tt}. \quad \text{(A2-11)}$$

Assumptions I and II correspond to Eqs. (A2-10) and (A2-11), confirming logical self-consistency.

Substituting Eq. (A2-10) into Eq. (A2-7):

$$-E + \frac{q}{M}\left(\frac{\partial S}{\partial q}\right)^2 + U(q - q_0) = 0. \quad \text{(A2-12)}$$

Using Eqs. (A2-2) and (A2-12), we construct a Lagrange functional ($L\ (q,\psi)$) using the properties of the conjugate

functions:

$$L(q,\psi) \equiv \varepsilon = (U-E)\psi^*\psi + \frac{B^2}{M} p \left(\frac{\partial \psi^*}{\partial q}\right)\left(\frac{\partial \psi}{\partial q}\right). \tag{A2-13}$$

The variation problem is:

$$\delta \int L\,(q,\psi) dq = 0. \tag{A2-14}$$

Applying the Euler-Lagrange equation (to ($\psi^*$)), we derive the time-independent GBI probability wave equation:

$$\frac{B^2}{M}\left(q\frac{d^2\psi}{dq^2} + \frac{d\psi}{dq}\right) + [E - U(q-q_0)]\psi = 0, \tag{A2-15}$$

and

$$\int |\psi(p)|^2\, dp = 1 \ \text{ or } \ \textstyle\sum_i |\psi_i(p_i)|^2 = 1 \tag{A2-16}$$

Here, the normalized function ($|\psi(q)|^2$) represents the cumulative operant frequency/probability at ($q$), satisfying the normalization condition in Eq. (A2-16). Eq. (A2-15) is the GBI probability wave equation.

## Supplementary Material Information

Supplementary Material Information includes 4 Chinese stock market datasets available online with this manuscript.

## Ethical compliance

All procedures involving human participants were conducted in accordance with institutional and/or national research committee standards and the 1964 Helsinki Declaration and its later amendments or comparable ethical standards.

## Author contributions statement

Shi contributed half of this work, while the remaining authors contributed equally to the project. Specifically

Conceptualization: LS, HL, XG, WZ

Methodology: LS

Investigation: LS, HL and XG

Supervision: LS, HL, XG, and JO

Writing - original draft: LS, HL

Writing - review & editing: LS, HL, XG, WZ, and JO

Resource: LS and XG

All authors reviewed the manuscript.

## Additional information

All authors reviewed the manuscript.